# A ROBUST AND ADAPTABLE METHOD FOR FACE DETECTION BASED ON COLOR PROBABILISTIC ESTIMATION TECHNIQUE

Reza Azad[1], Fatemeh Davami[2]

[1]IEEE Member, Department of Electrical and Computer Engineering, the SRTTU, Tehran, IRAN
E-mail: rezazad68@gmail.com

[2]Department of Electrical Engineering, Firoozabad Branch, Meymand Center, Islamic Azad University, Meymand, IRAN
E-mail: fatemeh.davami@gmail.com

***Abstract:*** *Human face perception is currently an active research area in the computer vision community. Skin detection is one of the most important and primary stages for this purpose. So far, many approaches are proposed to done this case. Near all of these methods have tried to find best match intensity distribution with skin pixels based on popular color spaces such as RGB, HSI or YCBCR. Results show that these methods cannot provide an accurate approach for every kind of skin. In this paper, an approach is proposed to solve this problem using a color probabilistic estimation technique. This approach is including two stages. In the first one, the skin intensity distribution is estimated using some train photos of pure skin, and at the second stage, the skin pixels are detected using Gaussian model and optimal threshold tuning. Then from the skin region facial features have been extracted to get the face from the skin region. In the results section, the proposed approach is applied on FEI database and the accuracy rate reached 99.25%. The proposed approach can be used for all kinds of skin using train stage which is the main advantage among the other advantages, such as Low noise sensitivity and low computational complexity.*

***Keywords:*** *Face Detection, Image Processing, Threshold tuning, Gaussian Model feature extraction*

## I. INTRODUCTION

Human face perception is currently an active research area in the computer vision community. Human face localization and detection is often the first step in applications such as video surveillance, human computer interface, face recognition and image database management. Locating and tracking human faces is a prerequisite for face recognition and/or facial expressions analysis, although it is often assumed that a normalized face image is available. In order to locate a human face, the system needs to capture an image using a camera and a frame-grabber to process the image, search the image for important features and then use these features to determine the location of the face. One the most important stages which should be used to face detection, is skin detection. In this respect, many approaches have proposed to detect skin which provide high detection rate. Some of them are used RGB color space to solve this problem such as [1], [2] and [3]. Some of the researchers used other color spaces such as $YC_bC_r$ [4], HSV [5], CIE LUV [6], and Farnsworth UCS [7]. Color is an important feature of human faces. Using skin-color as a feature for tracking a face has several advantages. Color processing is much faster than processing other facial features. Under certain lighting conditions, color is orientation invariant. This property makes motion estimation much easier because only a translation model is needed for motion estimation. However, color is not a physical phenomenon; it is a perceptual phenomenon that is related to the spectral characteristics of electromagnetic radiation in the visible wavelengths striking the retina. Tracking human faces using color as a feature has several problems like the color representation of a face obtained by a camera is influenced by many factors (ambient light, object movement, etc.), different cameras produce significantly different color values even for the same person under the same lighting conditions and skin color differs from person to person.

In order to use color as a feature for face tracking, we have to solve these problems. For solving these problems such as different kind of skin, lighting condition and etc. in this paper an approach is proposed based on probabilistic estimation technique. In this approach which contains two stages, first of all, some train images which include just pure skin pixels of slightly kind of skin is provided. Next, mean and standard deviation of these pixels are computed for each RGB channels individually. In the test stage, for all of the test pixels, the probabilistic P (RGBTrain|Skin) is computed using Gaussian model in

which one of the RGB channels individually. Finally, using an accurate and separable threshold can detect skin pixels in test images. Then from the skin region facial features have been extracted to get the face from the skin region. In the result part, proposed approach is applied on test images and the accuracy rate is computed. Some of the proposed approach advantages are: 1) Adaptability to most kinds of skin by using a train stage;2) Low complexity in computation and time;3) Low sensitivity to illumination by using tunable threshold.

In the next section we have explained the color spaces used to classify the skin color and well-known algorithms based on RGB, YCbCr and HSI color spaces. Proposed method has been detailed in third section. Implementation and Experimental results have been explained in fourth and fifth section respectively and sixth section is conclusion and future work.

## II. COLOR MODELS FOR SKIN COLOR CLASSIFICATION

The study on skin color classification has gained increasing attention in recent years due to the active research in content-based image representation. For instance, the ability to locate image object as a face can be exploited for image coding, editing, indexing or other user interactivity purposes. Moreover, face localization also provides a good stepping stone in facial expression studies. It would be fair to say that the most popular algorithm to face localization is the use of color information, whereby estimating areas with skin color is often the first vital step of such strategy. Hence, skin color classification has become an important task. Much of the research in skin color based face localization and detection is based on RGB, YCbCr and HSI color spaces. In this section the color spaces and some basic algorithm are being described.

### *A. RGB Color Space*

The RGB color space consists of the three additive primaries: red, green and blue. Spectral components of these colors combine additively to produce a resultant color. The RGB model is represented by a 3-dimensional cube with red green and blue at the corners on each axis (Figure 1). Black is at the origin. White is at the opposite end of the cube. The gray scale follows the line from black to white. In a 24-bit color graphics system with 8 bits per color channel, red is (255, 0, 0). On the color cube, it is (1, 0, 0). The RGB model simplifies the design of computer graphics systems but is not ideal forall applications. The red, green and blue color components are highly correlated. This makes it difficult to execute some image processing algorithms. Many processing techniques, such as histogram equalization, work on the intensity component of an image only.

Crowley and Coutaz [8] said one of the simplest algorithms for detecting skin pixels is to use skin color algorithm. The perceived human color varies as a function of the relative direction to the illumination. The pixels for skin region can be detected using a normalized color histogram, and can be further normalized for changes in intensity on dividing by luminance. And thus converted an [R, G, B] vector is converted into an [r, g] vector of normalized color which provides a fast means of skin detection. This gives the skin color region which localizes face. As in [8], the output is a face detected image which is from the skin region. This algorithm fails when there are some more skin region like legs, arms, etc.

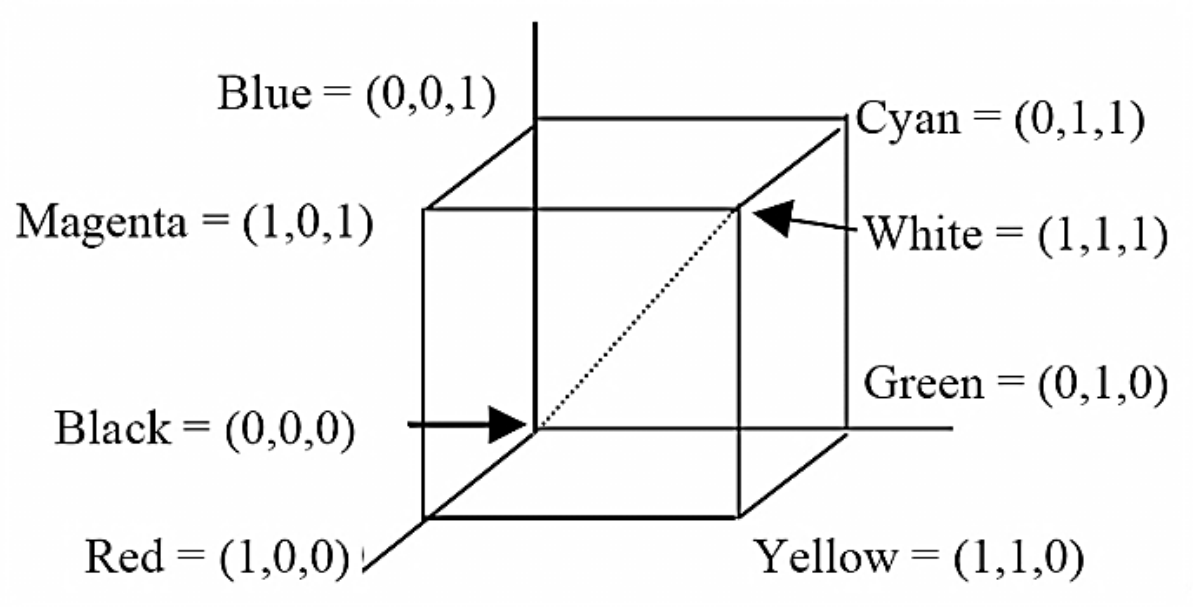

*Figure 1: RGB color*

### *B. YCbCr Color Space*

YCbCr color space has been defined in response to increasing demands for digital algorithms in handling video information, and has since become a widely used model in a digital video. It belongs to the family of television transmission color spaces. The family includes others such as YUV and YIQ. YCbCr is a digital color system, while YUV and YIQ are analog spaces for the respective PAL and NTSC systems. These color spaces separate RGB (Red-Green-Blue) into luminance and chrominance information and are useful in compression applications however the specification of colors is somewhat unintuitive. The Recommendation 601 specifies 8 bit (i.e. 0 to 255) coding of YCbCr, whereby the luminance component Y has an excursion of 219 and an offset of +16. This coding places black at code 16 and white at code 235. In doing so, it reserves the extremes of the range for signal processing footroom and headroom. On the other hand, the chrominance components Cb and Cr have excursions of +112 and offset of +128, producing a range from 16 to 240 inclusively.

Cahi, D. and Ngan have implemented a skin color classification algorithm [9] with color statistics gathered from YCbCr color space. Studies have found that pixels belonging to skin region exhibit similar Cb and Cr values. Furthermore, it has been shown that skin color model based on the Cb and Cr values can provide good coverage of different human races. The thresholds be chosen as [Cr1, Cr2] and [Cb1, Cb2], a pixel is classified to have skin tone if the values [Cr, Cb] fall within the thresholds The skin color distribution gives the face portion in the color image. This algorithm is also having the constraint that the image should be having only face as the skin region.

### *C. HSI Color Space*

Since hue, saturation and intensity are three properties used to describe color, it seems logical that there be a corresponding color model, HSI. When using the HSI color space, you don't need to know what percentage of blue or green is required to produce a color. You simply adjust the hue to get the color you wish. To change a deep red to pink, adjust the saturation. To make it darker or lighter, alter the intensity. Many applications use the HSI color model. Machine vision uses HSI color space in identifying the color of different objects. Image processing applications such as histogram operations, intensity transformations and convolutions operate only on an intensity image. These operations are performed with much ease on an image in the HSI color space. For the HSI being modeled with cylindrical coordinates, see Figure 2. The hue (H) is represented as the angle 0, varying from 0o to 360o. Saturation (S) corresponds to the radius, varying from 0 to 1. Intensity (I) varies along the z axis with 0 being black and 1 being white. When S = 0, color is a gray value of intensity 1. When S = 1, color is on the boundary of top cone base. The greater the saturation, the farther the color is from white/gray/black (depending on the intensity). Adjusting the hue will vary the color from red at 0o, through green at 120o, blue at 240o, and back to red at 360o. When I = 0, the color is black and therefore H is undefined. When S = 0, the color is gray scale. H is also undefined in this case. By adjusting I, a color can be made darker or lighter. By maintaining S = 1 and adjusting I, shades of that color are created.

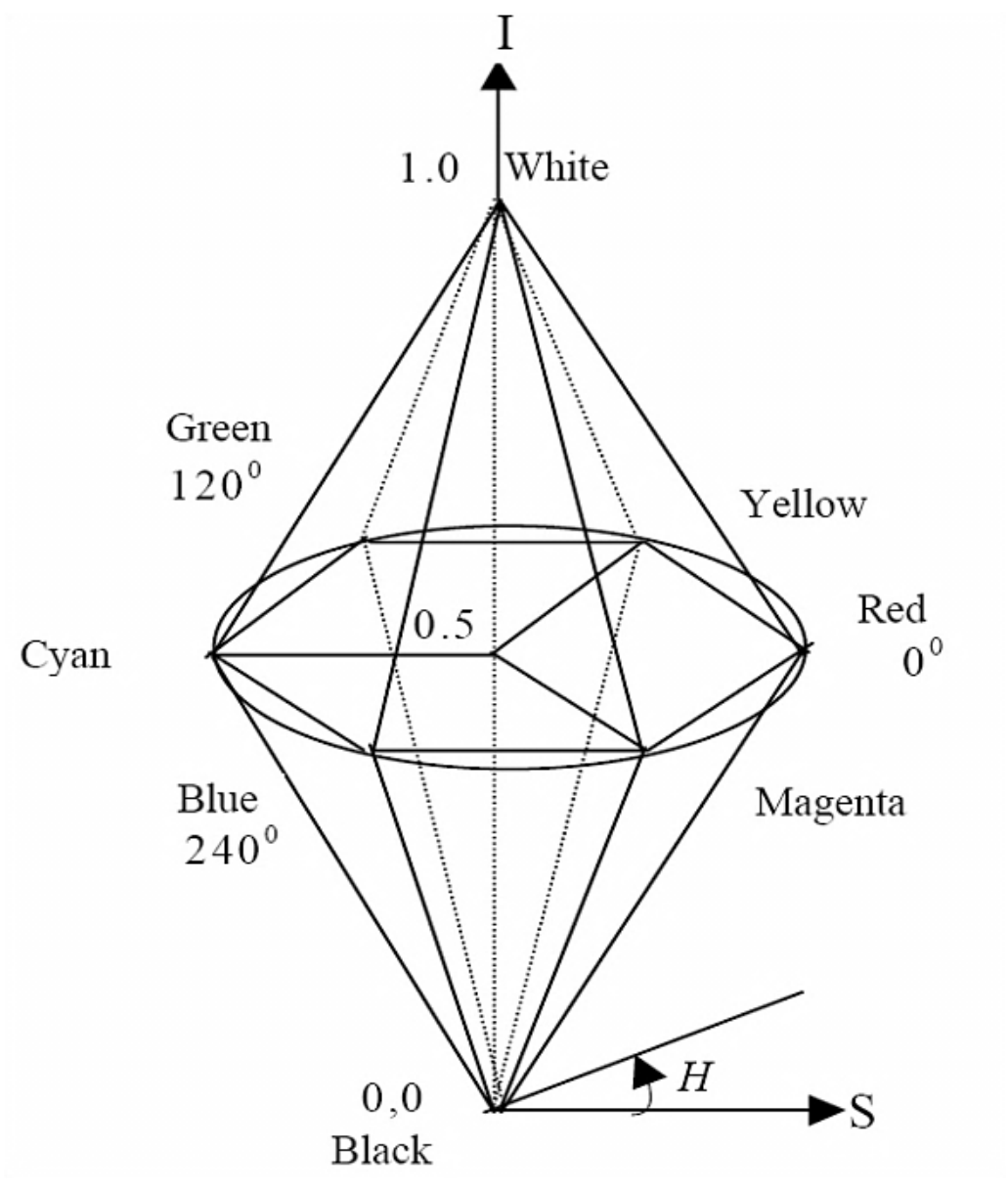

*Figure 2: Double cone model of HSI color space*

Kjeldson and Kender defined a color predicate in HSV color space to separate skin regions from background [10]. Skin color classification in HSI color space is the same as YCbCr color space but here the responsible values are hue (H) and saturation (S). Similar to above the threshold can be chosen as [H1, S1] and [H2, S2], and a pixel is classified to have skin tone if the values [H,S] fall within the threshold and this distribution gives the localized face image. Similar to above two algorithm this algorithm is also having the same constraint.

## III. PROPOSED METHOD

Proposed method described in the next subsections.

### *A. Estimation of Skin distribution*

In this approach, it is too important to estimate the skin distribution accurately. According to mentions which are discussed in the introduction part, some train images which contains just pure skin pixels, should be provided. In [11], Akhloufi et al. are proved that study RGB channels individually can provide discriminant features. Akhloufi et al. are computed statistical features such as mean, entropy, energy, standard deviation and etc. from Red, Green and blue histograms individually to classification the textures. In this respect, in this paper, in order to estimate the skin pixels distribution, it’s enough to compute statistical features like mean and standard deviation of skin pixels in each channel. It is shown in (1), and (2) for Red channel.

$$\mathrm{Mean(R)} = {}^{1}/_{\mathrm{n}\times\mathrm{m}} \sum_{\substack{0\le \mathrm{i} \le \mathrm{m}\\ 0<j<n}} \mathrm{P(i,j,1)} \qquad (1)$$

$$STD(R) = \sqrt{{}^{1}/_{n\times m} \sum_{\substack{0\le i\le m\\ 0<j<n}} (P(i,j,1) - mean(R))^2} \qquad (2)$$

Where, P(i, j, 1) means the intensity of pixel in $i_{th}$ row and $j_{th}$ column in red channel. Also, m and n are the size of train image. By using a similar way, the mean and STD in green and blue channels can be computed. Now, the feature vector, like F which is shown in (3) can be provided as a good identification of skin and shows distribution of the skin pixels in each RGB channel.

$$F = (Mean_R, Mean_G, Mean_B, STD_R, STD_G, STD_B) \qquad (3)$$

### *B. Gaussian Model for skin detection*

In test stage, the aim is to estimate the probability of each test pixels using skin distribution. In this respect, the Gaussian model can be used to compute P ($RGB_{Test}$| Skin). In order to estimate P ($RGB_{Test}$|Skin), by assuming independency between RGB channels, we have (4).

$$P(RGB_{Test}|\,Skin) = P(R_{Test}|Skin) \times P(G_{Test}|Skin) \times P(B_{Test}|Skin) \quad (4)$$

Now, Gaussian model can be used to compute the probability estimation between RGB channels and Skin distribution. (5), shows these algorithms. Where N is the Gaussian distribution and R, G and B are test pixels intensities in each channel.

$$P(Test \mid skin) = \{N(R_{Test}, Red\text{–}mean_{Train}, Red\text{-}std_{Train}) \times N(G_{Test}, Green_mean_{Train}, Green\text{-}std_{Train}) \times N(B_{Test}, Blue_mean_{Train}, Blue\text{-}std_{Train})\} \quad (5)$$

Also, (6) shows the Gaussian distribution for input x. The STD is the square root of an unbiased estimator of the variance of the population from which X is drawn, as long as X consists of independent, identically distributed samples.

$$F(x, mean, STD) = exp\left\{{-0.5 \times (x - mean)^2}/{STD}\right\} \quad (6)$$

## C. THRESHOLD TUNING

In order to increase the quality of the proposed skin detection approach a preprocessing stage is needed. Preprocessing should normal the histogram of intensities and decreases the illumination. In this respect, histogram equalization technique is used. Histogram equalization transforms an image with an arbitrary histogram to one with a flat histogram.

Using a threshold, after computing probability of all test pixels, can solve the skin detection. In this respect, calculating an accurate and severable threshold is necessary. In order to find tuned threshold, it’s easy to compute the probability of all train pixels P $(RGB_{Train}|Skin)$ using an algorithm like which is proposed in section B. Next, it’s explicit that the lowest probability is a good and optimal threshold for skin detection. (7), shows this tuning method.

$$\begin{aligned} Threshold &= Min\{P(Train \mid Skin) \text{ For Each Train pixels}\} \\ &= Min\{N(R_{Train}, Red\text{-}mean_{Train}, Red\text{-}std_{Train}) \times N(G_{Train}, Green_mean_{Train}, Green\text{-}std_{Train}) \times N(B_{Train}, Blue_mean_{Train}, Blue\text{-}std_{Train}) \text{ For Each } RGB_{Train} \text{ pixels}\} \end{aligned} \quad (7)$$

Where N is the Gaussian distribution and R, G and B are train pixels intensities in each channel. For every kinds of skin, this tuning method can use. Also it will be provide the ability of skin kinds classification in common images. In Figure 3, skin detection is done based on some various thresholds and then they are compared by tuned optimal threshold.

## D. Face Detection

After getting the skin region, facial features such as eyes, ears and mouth are extracted. After thresholding, opening and closing operations are performed to remove noise. These are the morphological operations, opening operation is erosion followed by dilation to remove noise and closing operation is dilation followed by erosion which is done to remove holes. Now we extract the eyes, ears and mouth from the binary image by considering the threshold for areas which are darker in the mouth than a given threshold.

A triangle drawn with the two eyes and a mouth as the three points in case of a frontal face then we see that we get an isosceles triangle (i j k) in which the Euclidean distance between two eyes is about 90-110% of the Euclidean distance between the center of the right/left eye and the mouth. Figure 4 shows sample of this isosceles triangle. After getting the triangle, it is easy to get the coordinates of the four corner points that form the potential facial region. Since the real facial region should cover the eyebrows, two eyes, mouth and some area below the mouth, the coordinates can be calculated as follows: Assume that (Xi, Yi), (Xj,Yj) and (Xk, Yk) are the three center points of blocks i, j, and k, that form an isosceles triangle.(X1, Y1), (X2, Y2), (X3, Y3) and (X4, Y4) are the four corner points of the face region. X1 and X4locate at the same coordinate of (Xi − 1/3D(i, k));X2 and X3 locate at the same coordinate of (Xk +1/3D(i, k)); Y1 and Y2 locate at the same coordinates of (Yi + 1/3D(i, k)); Y3 and Y4 locate at the same coordinates of (Yj − 1/3D(i, k)); where D(i, k)is the Euclidean distance between the centers of block i (right eye) and block k (left eye).

$$X1 = X4 = Xi - 1/3D(i, k) \quad (8)$$
$$X2 = X3 = Xk + 1/3D(i, k) \quad (9)$$
$$Y1 = Y2 = Yi + 1/3D(i, k) \quad (10)$$
$$Y3 = Y4 = Yj - 1/3D(i, k) \quad (11)$$

This is done for a frontal face; similarly for the side (left/right) view. For the side view a right triangle has obtained with the characteristic that the Euclidean distance of line i k is equal to 2 times of the Euclidean distance of line j k, and the Euclidean distance of line i j is equal to 1.732 times of the Euclidean distance of line j k. The 4 rules to get the face boundary for right side view are as follows

$$X1 = X4 = Xi - 1/6D(i, j) \quad (12)$$
$$X2 = X3 = Xi + 1.2D(i, j) \quad (13)$$
$$Y1 = Y2 = Yi + 1/4D(i, j) \quad (14)$$
$$Y3 = Y4 = Yi - 1.0D(i, j) \quad (15)$$

Similarly, the 4 rules to get the face boundary forleft side view are:

$$X1 = X4 = Xj - 1/6D(j, k) \quad (16)$$
$$X2 = X3 = Xj + 1.2D(j, k) \quad (17)$$
$$Y1 = Y2 = Yj + 1/4D(j, k) \quad (18)$$
$$Y3 = Y4 = Yj - 1.0D(j, k) \quad (19)$$

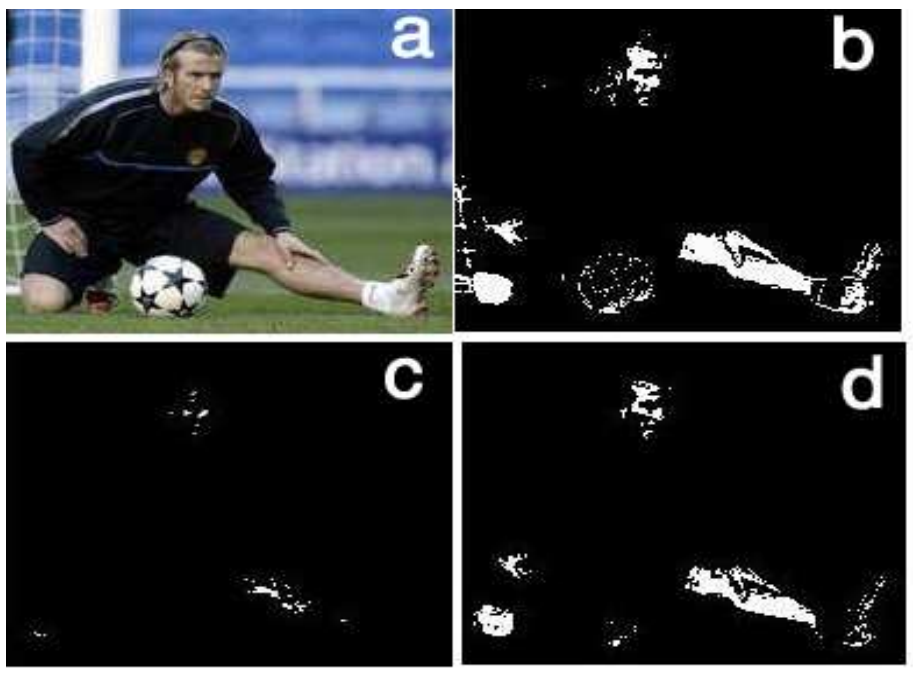

*Figure 3: (a) Original image (b) Threshold less than optimal*

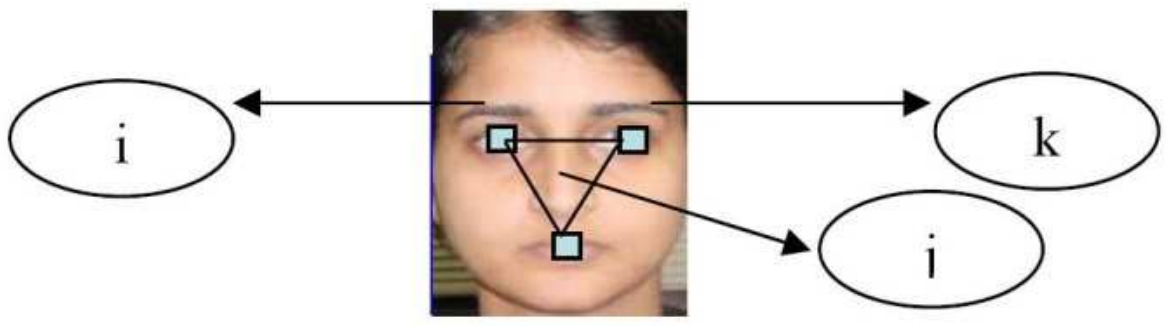

*Figure 4: Three points (i, j, and k) satisfy the matching*

## IV. IMPLEMENTATION

Our suggestive method have been done on Intel Core i3-2330M CPU, 2.20 GHz with 2 GB RAM under Matlab environment. The aim of this paper was to propose an accurate approach to detect face in images. According to this aim, an approach is proposed based on probabilistic estimation techniques and using a train stage. In the train stage, mean and standard deviation of the pure skin pixels are computed in each channel individually. Next, by using a Gaussian model of these, the probability of test pixels is estimated. Finally, the discriminative threshold which can provide a good severability between skin and non-skin pixels is tuned using a robust tuning algorithm. Figure 5 shows the face of worked systems for skin detection.

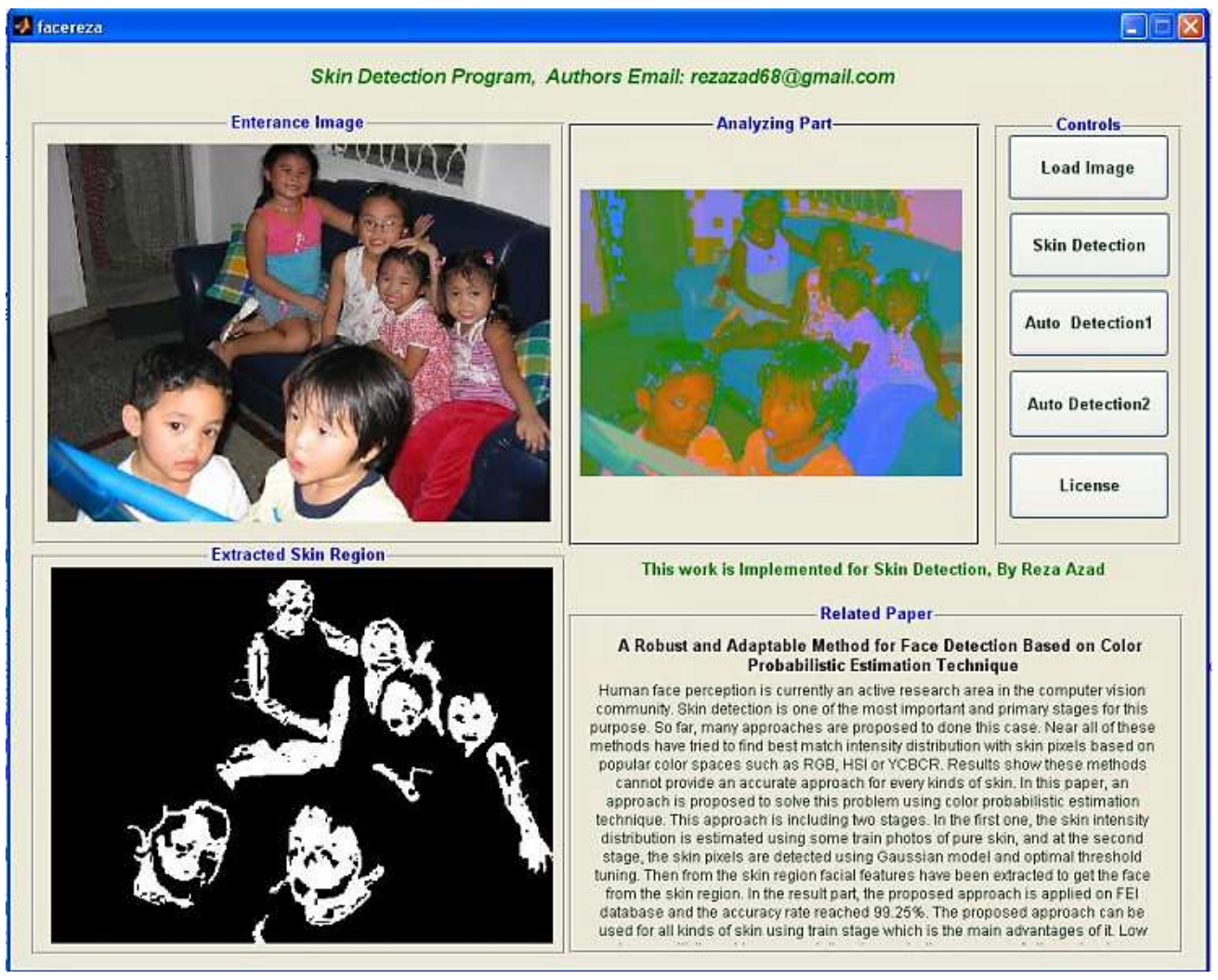

*Figure 5: Skin detection Part by Matlab environment*

Then from the skin region facial features have been extracted to get the face from the skin region. Figure 6 shows the face of worked systems for face detection.

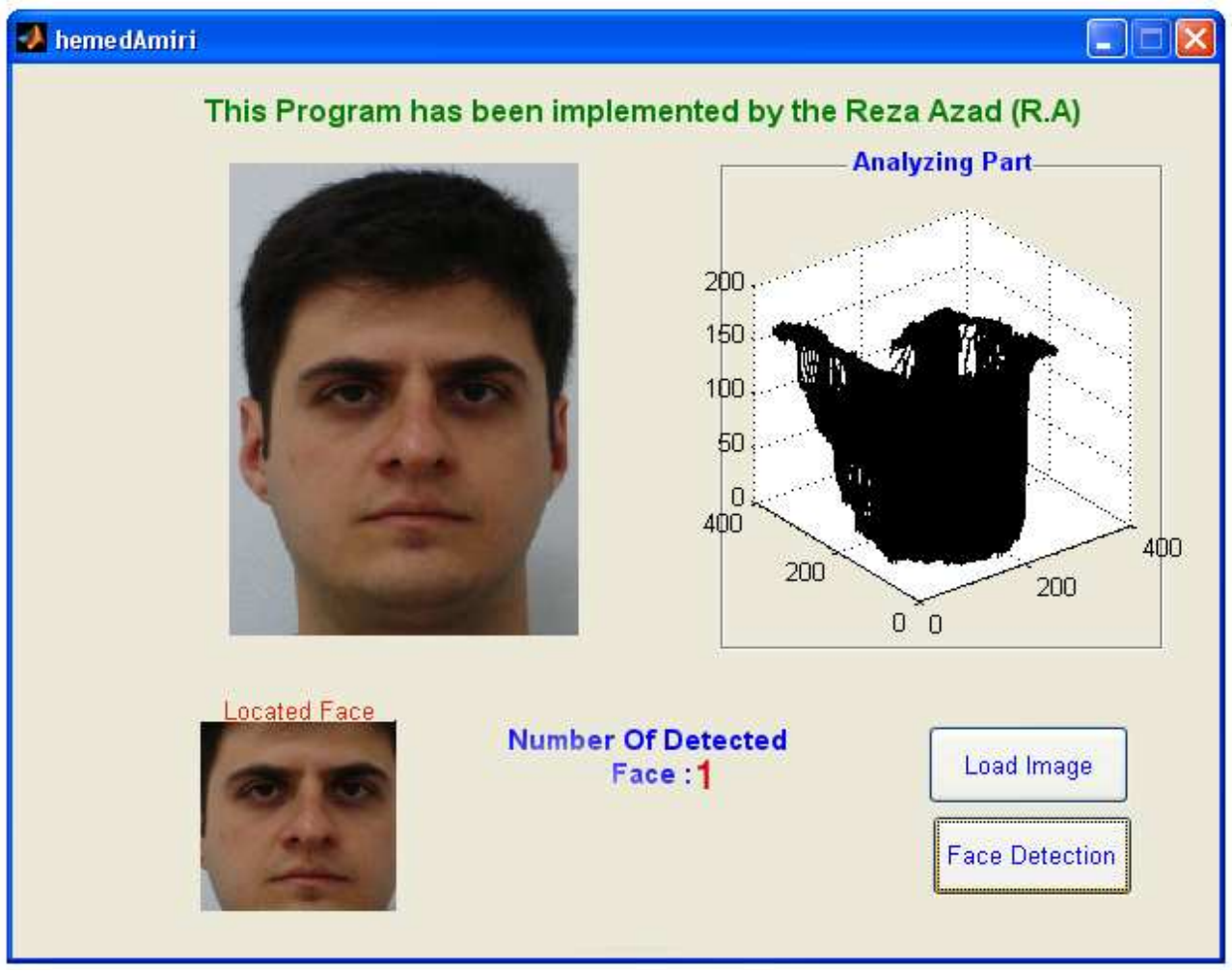

*Figure 6: Face detection system by Matlab*

## V. RESULT

In this study, for experimental analysis, we considered a FEI database for face detection. The FEI face database is a Brazilian face database that contains a set of face images taken between June 2005 and March 2006 at the Artificial Intelligence Laboratory of FEI in São Bernardo do Campo, São Paulo, Brazil. The results are as presented in table I.

*TABLE I: Comparison of different algorithms*

| Number of Images | FEI color image database | | |
|---|---|---|---|
| | **Method** | **Successful Face Localization** | **Localization Rate** |
| 400 | [12] | 394 | 98.5% |
| | Proposed Method | 397 | 99.25% |

Further we tasted our method for skin detection on the complex background images that mentioned in [13] for showing high accuracy of our method. The results are as presented in table II.

*TABLE II: Detection Rate on Complex Background Database*

| Number of Images | Complex Background database | | |
|---|---|---|---|
| | **Method** | **Successful Skin Localization** | **Detection Rate** |
| 100 | [13] | 100 | 95.40 ± 0.31 |
| | Proposed Method | 100 | 98.10 ± 0.19 |

High detection rate shows the quality of proposed approach to use in every applications, which are needed a skin detection stage. Low complexity in computation and time are some of other advantages of the proposed approach. Using a train stage is provide, various skin kinds detection ability. Also proposed approach can be used for skin kind's classification in complex images also. Complex images mean the images which are included two or more kinds of skins. Further, some other methods such as wavelet filtering, histogram analysis and Gabor filter are applied on test images, and the results are compared by proposed approach in terms of accuracy rate, which is mentioned in [14] are detailed in Table III. Following equation shows accuracy rate:

$$Accuracy = 100 - (False\ Detection\ Rate + False\ Dismissal\ Rate) \quad (20)$$

*TABLE III: Accuracy rates by using approaches*

| Approach | Detection rate |
|---|---|
| Wavelet | 97.64 ± 0.5 |
| Gabor | 94.73± 0.3 |
| Histogram Analysis | 98.20 ± 0.7 |
| Proposed Approach | 99.18 ± 0.2 |

Figure 7 Shows the sample of images from complex background database that the system be able to recognize them.

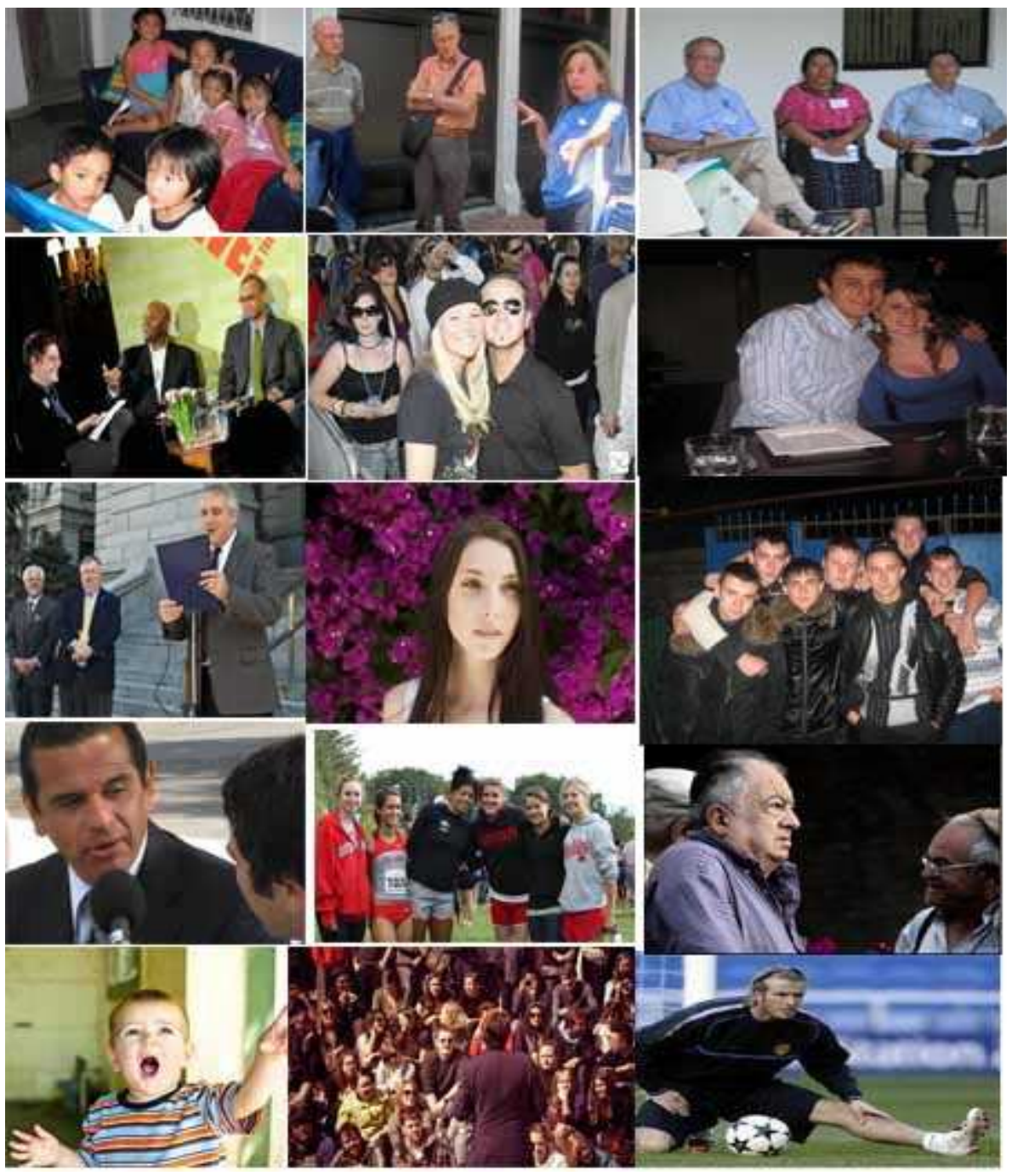

*Figure 7: Sample of complex background database images*

## VI. CONCLUSION AND FUTURE WORK

In this paper we propose an approach based on probabilistic estimation techniques for face detection in still images. First, in the train stage, mean and standard deviation of the pure skin pixels are computed in each channel individually. Next, by using a Gaussian model of these, the probability of test pixels is estimated. Finally, the discriminative threshold which can provide a good severability between skin and non-skin pixels is tuned using a robust tuning algorithm. Then from the skin region facial features have been extracted to get the face from the skin region. The result part is proved the quality of proposed approach to detect face pixels in terms of skin kinds. We tested our method on FEI and complex background database and the accuracy rate reached 99.25 and 98.10 ± 0.19respectively.

This work is implemented for still images, for future work we have planned to extend it for face detection in video stream. Also occasionally accuracy decreased when the background pixels have more similarity to skin pixels, for solving this problem in the future work we will express the method based on background subtracting stage to resolve this problem.